%% file: main.tex
\documentclass[twoside,11pt]{article}
\usepackage{amsfonts,amsmath}
\usepackage{jmlr2e}
\usepackage{amsfonts}
\usepackage{mathtools}
\usepackage{microtype}
\usepackage{graphicx}
\usepackage{subcaption}     
\usepackage{booktabs}       
\usepackage{hyperref}
\usepackage{adjustbox}      
\usepackage{todonotes}
\usepackage{xcolor}
 \usepackage{enumitem}

\jmlrheading{x}{x}{x}{x}{x}{xx}{All}

\begin{document}

\title{Optimal Transport Tools (OTT): \\A JAX Toolbox for all things Wasserstein}

\author{
        \name Marco Cuturi \email cuturi@google.com \\
       \addr Google, 8 rue de Londres, Paris, France
       \AND
       \name Laetitia Meng-Papaxanthos \email lpapaxanthos@google.com \\
       \addr Google, Brandschenkestrasse 110, Zürich, Switzerland
       \AND
       \name Yingtao Tian \email alantian@google.com \\
       \addr Google, Shibuya 3-21-3, Tokyo, Japan
       \AND 
       \name Charlotte Bunne \email bunnec@ethz.ch \\
       \addr ETH Z\"urich
       \AND
       \name Geoff Davis \email geoffd@google.com \\
       \addr Google, Mountain View, CA, United States
       \AND
       \name Olivier Teboul \email oliviert@google.com \\
       \addr Google, 8 rue de Londres, Paris, France
       }

\editor{xxx}

\maketitle

\begin{abstract}
Optimal transport tools (OTT-JAX) is a python toolbox that can solve optimal transport problems between point clouds and histograms. The toolbox builds on various JAX features, such as automatic and custom reverse mode differentiation, vectorization, just-in-time compilation and accelerators support. 
The toolbox covers elementary computations, such as the resolution of the regularized OT problem, and more advanced extensions, such as barycenters, Gromov-Wasserstein, low-rank solvers, estimation of convex maps, differentiable generalizations of quantiles and ranks, and approximate OT between Gaussian mixtures. The toolbox code is available at \texttt{https://github.com/ott-jax/ott}
\end{abstract}


\input{sections/intro}
\input{sections/implementation}
\input{sections/application}
\bibliography{all.bib}

\end{document}

%% file: sections/intro.tex
\section{Background and Motivation: Optimal Transport in a Nutshell}
Many areas in science and engineering have to deal with tasks that involve pairing points across two datasets. A minimal working example for such tasks arises when a family of points $(x_1, \dots, x_n)$, taken in a space $\mathcal{X}$ has to be matched (registered, assigned, aligned) to another family $(y_1, \dots, y_n)$ of the same size, taken in a possibly different space $\mathcal{Y}$. A \emph{matching} between these two families is a bijective map that assigns to each point $x_i$ another point $y_j$, and is usually encoded as a \textit{permutation} $\sigma$. With that convention $x_i$ would be assigned to $y_{\sigma_i}$, the $\sigma_i$-th element in the second family. To define what constitutes a ``good'' assignment, a practitioner defines an objective $\mathcal{E}$ on permutations. The simplest one is that which is parameterized by a \textit{ground} cost function $c$ between points, $c:\mathcal{X}\times\mathcal{Y}\rightarrow \mathbb{R}$, to define the objective
$\mathcal{E}_1(\sigma) = \frac{1}{n}\sum_{i=1}^n c(x_i, y_{\sigma_i})$.

Opening a window to more complex transport problems, situations may arise when the user may not have sufficient prior information to define the ground cost function $c$. This arises notably when $\mathcal{X}\ne\mathcal{Y}$ (when $\mathcal{X}=\mathcal{Y}$, $c$ is usually easier to pick, e.g., a distance). In such settings, and because it is comparatively easier to define two cost functions $c_{\mathcal{X}}$ and $c_{\mathcal{Y}}$ defined respectively on $\mathcal{X}\times \mathcal{X}$ and $\mathcal{Y}\times\mathcal{Y}$ (for instance distances), one can consider instead the following objective, that uses a discrepancy function $\Delta$ between reals, $\Delta:\mathbb{R}\times\mathbb{R}\mapsto\mathbb{R}$,
$$
\mathcal{E}_2(\sigma) = \frac{1}{n^2}\sum_{i,i'=1}^n \Delta\left(\,c_{\mathcal{X}}(x_i, x_{i'}), c_{\mathcal{Y}}(y_{\sigma_i}, y_{\sigma_{i'}}\,)\right)\,.
$$
Although energy $\mathcal{E}_2$ is slightly harder to parse than $\mathcal{E}_1$, its role is to favor permutations $\sigma$ that are as close as possible to an isometry: ideally, if $x_i$ is assigned to $y_{\sigma_i}$ and $x_{i'}$ to $y_{\sigma_{i'}}$, one hopes that $c_{\mathcal{X}}(x_i, x_{i'})$ has a value that is close to $c_{\mathcal{Y}}(y_{\sigma_i}, y_{\sigma_{i'}})$.
Several issues arise when trying to optimize $\mathcal{E}_1$ or $\mathcal{E}_2$. The first one lies in the challenge brought forward by optimizing over permutations. This is typically solved by using a relaxation, namely by representing permutations as $n\times n$ matrices with binary entries, with the constraint that each of their rows and columns has exactly one and one only non-zero entry equal to 1. From there, the binary constraint can be relaxed using \citeauthor{birkhoff}'s polytope \citeyearpar{birkhoff}, to focus on bistochastic matrices. The second issue arises when the size of the two families are not the same, and/or weights are associated to the observations, in order to compare two discrete measures $\mu=\sum_{i=1}^n a_i \delta_{x_i}$ and $\nu=\sum_{j=1}^m b_j \delta_{y_j}$. The set of matrices used to represent relaxed permutations is called the transportation polytope, and reads:
$$U(a,b) = \{P \in\mathbb{R}^{n\times m}_+\,, P\mathbf{1}_m = a, P^T\mathbf{1}_n = b\}, \text{ assuming } a^T\mathbf{1}_n = b^T\mathbf{1}_m=1.$$
The focus of the toolbox will be to solve two problems in $U(a,b)$, optimizing the linear and quadratic objectives $\mathcal{E}_1, \mathcal{E}_2$, reformulated to matching a transportation matrix variable $P$:

\begin{align}
    \mathcal{L}_c(\mu, \nu) &= \min_{P \in U(a,b)} \sum_{i,j=1,1}^{n,m} c(x_i, y_j)\,P_{ij}\,.\label{eq:lin}\\
    \mathcal{Q}_{c_\mathcal{X}, c_\mathcal{Y}}(\mu, \nu) &= \min_{P \in U(a,b)} \sum_{i,i',j,j'=1,1,1,1}^{n,n,m,m}  \Delta\left(\,c_{\mathcal{X}}(x_i, x_{i'}), c_{\mathcal{Y}}(y_j, y_{j'})\,\right)\,P_{ij}P_{i'j'}\,.\label{eq:quad}
\end{align}

Problems \eqref{eq:lin} and \eqref{eq:quad} can be easily extended to a continuous setting. This requires defining joint couplings with marginals $\mu, \nu$, and reformulate the same objectives.
\begin{align}
    \Pi(\mu,\nu) &= \{\pi \in\mathcal{P}(\mathcal{X}\times\mathcal{Y}), \forall A,B\subset \mathcal{X}\times\mathcal{Y}, \pi(A\times \mathcal{Y})=\mu(A), \pi(\mathcal{X}\times B)=\nu(B)\}\\
    \mathcal{L}_c(\mu, \nu) &= \min_{\pi \in \Pi(\mu,\nu)} \iint_{\mathcal{X}\times\mathcal{Y}} c(x,y)\, \textrm{d}\pi(x,y)\label{eq:linc}\\
    \mathcal{Q}_{c_\mathcal{X}, c_\mathcal{Y}}(\mu, \nu) &= \min_{\pi \in \Pi(\mu,\nu)} \iint_{\mathcal{X}^2}\iint_{\mathcal{Y}^2} \Delta\left(\,c_{\mathcal{X}}(x, x'), c_{\mathcal{Y}}(y, y')\,\right)\textrm{d}\pi(x,y)\textrm{d}\pi(\textrm x', y')\,.\label{eq:quadc}
\end{align}
We overload notations here because instantiating~\eqref{eq:linc} and \eqref{eq:quadc} with discrete measures yields back \eqref{eq:lin} and \eqref{eq:quad}. The relaxation \eqref{eq:lin}, and its continuous extension, are attributed to \cite{Kantorovich42}, while \eqref{eq:quad} comes from~\cite{memoli-2011}. Solving these relaxed problems remains, however, an arduous task. We narrow down on three issues:

\begin{enumerate}
    \item \textbf{Scalability.} While \eqref{eq:lin} can be solved using a network-flow solver, known worst-case bounds are super-cubic~\citep{ahuja1988network}, e.g. $O(nm(n+m)\log(n+m))$. Problem \eqref{eq:quad} is NP-hard (notably for $\Delta(u,v)=(u-v)^2$), and usually solved heuristically through iterative linearization, to fall back on~\eqref{eq:lin}, having to pay repeatedly for that cubic price.
    
    \item \textbf{Curse of Dimensionality.} Most practitioners will likely aim at matching, ideally, densities $\mu, \nu$ as in ~\eqref{eq:linc} and \eqref{eq:quadc}, yet, because these densities are likely to be only reachable (notably in high-dimensions) through samples, computations must be carried out using empirical measures $\hat\mu_n$, $\hat\nu_n$. Computing $\mathcal{L}_c(\hat\mu_n,\hat\nu_n)$ exactly, as in~\eqref{eq:lin} to approach $\mathcal{L}_c(\mu,\nu)$, as in \eqref{eq:linc}, is doomed by the curse of dimensionality~\cite{fournier2015rate}. In essence, it is akin to wasting compute resources to overfit these samples. Although not as well understood, the curse of dimensionality is likely to be at least (if not more) problematic for $\mathcal{Q}_{c_\mathcal{X}, c_\mathcal{Y}}$.
    
    \item \textbf{Argmin Diferentiability.} While many practitioners might only focus on the value of $\mathcal{L}_c$ or $\mathcal{Q}_{c_\mathcal{X}, c_\mathcal{Y}}$, an increasing number of applications may focus instead on optimal $P^\star$ matrix solutions. While differentiating $\mathcal{L}_c$ or $\mathcal{Q}_{c_\mathcal{X}, c_\mathcal{Y}}$ w.r.t. any of its inputs can be computed using the envelope (a.k.a. Danskin) theorem, variations in $P^\star$ are far less smooth. For instance, the Jacobian $J_{x_i}P^\star$ of the optimal transport matrix w.r.t. an arbitrary input point $x_i$ is almost everywhere $0$ (the optimal match is usually not affected by an infinitesimal change).
\end{enumerate}

\textbf{The goal of OTT} is to solve variants of the original OT problems having these issues in mind. Entropic regularization~\citep{CuturiSinkhorn} is known to help with all three items, notably statistical aspects~\citep{genevay2018sample,mena2019statistical} and differentiability, using either unrolling~\citep{adams2011ranking,2016-bonneel-barycoord} or implicit differentiation
~\citep{luise2018differential,cuturi2020supervised}. The Low-Rank Sinkhorn approach~\citep{pmlr-v139-scetbon21a} addresses computational challenges further, by imposing a low-rank structure for $P^\star$, which can also be leveraged for the quadratic case~\citep{scetbon2021linear}.

%% file: sections/implementation.tex
\section{Implementation}
We review in this section the essential components of the toolbox.

\paragraph{\texttt{geometry} folder.}
The \texttt{Geometry} class and inherited classes that build on it can be thought as encapsulating the mathematical properties of cost matrices $[c(x_i, y_j)]_{ij}$, $[c_\mathcal{X}(x_i, x_{i'})]_{ii'}$ or $[c_\mathcal{Y}(y_j, y_{j'})]_{jj'}$ appearing in either or both Problem~\eqref{eq:lin} and Problem~\eqref{eq:quad}, with the aim of not having to store them as matrices $C$ in memory. For instance, \textit{(i)} when the cost is the squared-Euclidean norm of vectors in $\mathbb{R}^d$ (the scenario considered when instatiating a \texttt{PointCloud} geometry), it is known that $C$ has rank up to $d+2$, making an exact storage of all $n\times m$ distances useless; \textit{(ii)} when measures are supported in a grid that is the Cartesian product of $d$ univariate discretizations, the cost matrix $C$ would be theoretically of size $n^d \times n^d$, but applying $C$ or its kernel $e^{-C/\varepsilon}$ can be carried out in $O(dn^{d+1})$ operations.

\paragraph{\texttt{core} folder.}
The \texttt{core} section of the toolbox holds all of the crucial ingredients required to solve approximately both~\eqref{eq:lin} and, consequently,~\eqref{eq:quad}. 
\begin{itemize}[leftmargin=.5cm,itemsep=.01cm,topsep=0cm]
\item \textbf{Problem definitions}, \texttt{problems.py,quad\_problems.py}: \texttt{Problems} are classes that describe a task to be solved. These problems can be defined in various flavours, either linear, or, possibly, quadratic and even fused. They essentially contain a \texttt{Geometry} object with probability weights.

\item \textbf{Entropic and Low-Rank Sinkhorn Solvers}, \texttt{sinkhorn.py,sinkhorn\_lr.py}: Solvers to solve iteratively~\eqref{eq:lin}, using either a Sinkhorn approach or low-rank constraints.
\item \textbf{Entropic and Low-Rank Gromov Solvers}, \texttt{gromov\_wasserstein.py}: Solver class to address~\eqref{eq:quad}, by making iterated calls to linear problems instantiated above.
\item \textbf{Regularized Barycenters}, \texttt{discrete\_barycenter.py}: Computation of barycenters between discrete measures, assuming the support of the barycenter is fixed.
\item \textbf{Neural Potentials}, \texttt{icnn.py}: input-convex neural networks~\citep{amos2017input}.
\end{itemize}

\paragraph{\texttt{tools} folder.} 
\begin{itemize}[leftmargin=.5cm,itemsep=.01cm,topsep=0cm]
\item \textbf{Soft-sorting}, \texttt{soft\_sort.py}: implements ideas from \cite{NEURIPS2019_d8c24ca8,cuturi2020supervised} that highlight OTT's ability to differentiate through the optimal solutions $P^\star$ of~\eqref{eq:lin}.
\item \textbf{Gaussian Mixtures} folder computes a Wasserstein-like distance between Gaussian mixtures from \cite{delon2020}, which can be used to fit pairs of GMMs with a map between them.
\end{itemize}

%% file: sections/application.tex
\section{Applications}
A typical OTT pipeline involves instantiating \texttt{geom = PointCloud(x, y, cost=...)}, a geometry from points, where \texttt{x} and \texttt{y} are respectively $n\times d$ and $m\times d$ matrices, possibly endowed with a custom cost function. Specifying next two marginal probability vectors of size $n$ and $m$, \texttt{a} and \texttt{b}, one can define an OT problem, \texttt{prob=LinearProblem(geom, a, b))} to which a Sinkhorn solver, or a LR Sinkhorn solver can be applied, \texttt{out= Sinkhorn()(prob)} or \texttt{out= LRSinkhorn(rank=r)(prob)}. The output encapsulates the transport's properties, such as, naturally, an \texttt{out.matrix}.
Figure~\ref{fig:applications} shows several advanced applications of OTT-JAX that demonstrates new algorithms combined with differentiability. 
Details of a wide selection of advances applications, including that of those in Figure~\ref{fig:applications}, are in the toolbox's documentation as Jupyter Notebooks, which can easily run online without any setup using Colab. 
\input{figures/applications}

%% file: figures/applications.tex
\begin{figure}[!hbt]
    \centering
    \begin{tabular}{ccc}
    \includegraphics[width=.3\textwidth]{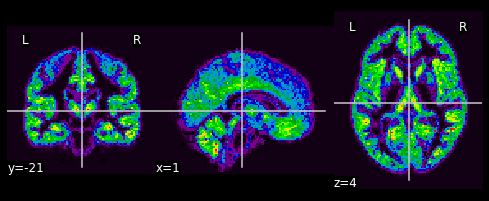}&
    \includegraphics[width=.3\textwidth]{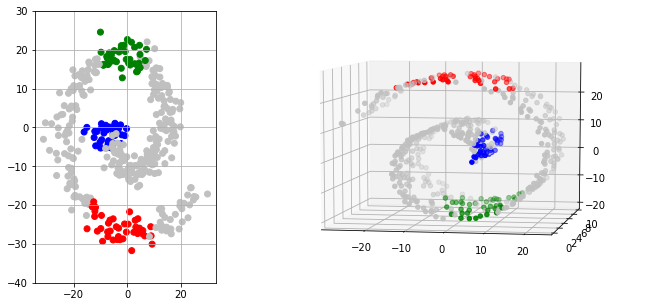}&    
    \includegraphics[width=.3\textwidth]{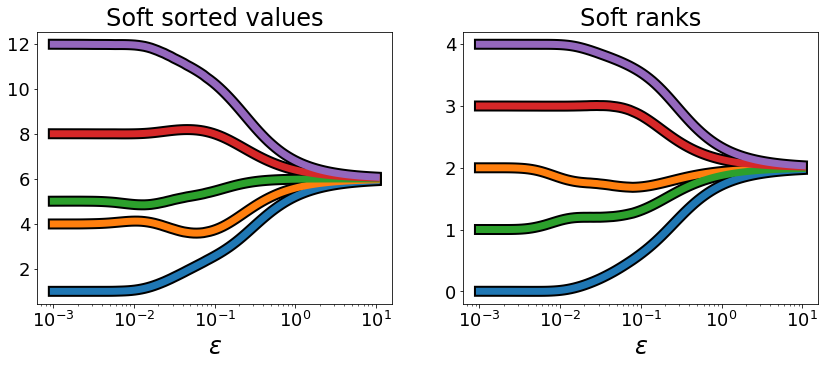}
    \end{tabular}
    \caption{Three advanced applications using OTT-JAX: \textit{(left)}  regularized $W_p^p$ iso-barycenter of Voxel-Based Morphometry on Oasis dataset, computed efficiently using grid geometry. \textit{(middle)} Gromov Wasserstein problem matching a spiral in 2 dimensions and a Swiss roll in 3 dimensions where associated points are in the same color.\textit{(right)} soft-sorted array $[1.0, 5.0, 4.0, 8.0, 12.0]$ with different $\epsilon$ controlling differentiability.
    }
    \label{fig:applications}
\end{figure}